\documentclass[times,final]{elsarticle}

\usepackage{lineno,hyperref}
\usepackage[english]{babel}
\usepackage{xcolor}
\usepackage{subfiles}
\usepackage{blindtext}
\usepackage[normalem]{ulem}
\graphicspath{{images/}{../images/}}

\usepackage{graphicx}
\usepackage{amssymb}

\usepackage{lineno}
\usepackage{amssymb}
\usepackage{mathtools}

\usepackage{amsmath}
\DeclareMathOperator*{\argmin}{argmin} 

\newcommand{\real}{{\mathbb R}}

\newcommand{\n}{{\mathbb N}}

\makeatletter
\def\ps@pprintTitle{%
   \let\@oddhead\@empty
   \let\@evenhead\@empty
   \let\@oddfoot\@empty
   \let\@evenfoot\@oddfoot
}
\makeatother
\begin{document}
\begin{frontmatter}

\title{Diagnosis of Pediatric Obstructive Sleep Apnea via Face Classification with Persistent Homology and Convolutional Neural Networks}



\author[uofaaddress]{Milad Kiaee}
\author[uofaaddress2]{Adam B Kashlak}
\author[inriaaddress]{Jisu Kim}

\author[uofaaddress]{ Giseon Heo \corref{mycorrespondingauthor}}
\cortext[mycorrespondingauthor]{Corresponding author}
\ead{gheo@ualberta.ca}

\address[uofaaddress]{School of Dentistry, University of Alberta}
\address[uofaaddress2]{Department of Mathematical and Statistical Sciences, University of Alberta}
\address[inriaaddress]{Inria Researh Centre}

\begin{abstract}
  Obstructive sleep apnea is a serious condition causing a litany of health problems especially in the pediatric population. However, this chronic condition can be treated if diagnosis is possible. 
  The gold standard for diagnosis is an overnight sleep study, which is often 
  unobtainable by many potentially suffering from this condition. Hence, we attempt to develop a fast non-invasive diagnostic tool by training a classifier on 2D and 3D facial images of a patient to recognize facial features associated with obstructive sleep apnea. In this comparative study, we consider both persistent homology and geometric shape analysis from the field of computational topology as well as convolutional neural networks, a powerful method from deep learning whose success in image and specifically facial recognition has already been demonstrated by computer scientists.
\end{abstract}


\end{frontmatter}


\section{Introduction}

Obstructive sleep apnea (OSA) is a chronic condition characterized by frequent episodes of upper airway collapse during sleep. Pediatric OSA is a serious health problem as even mild forms of untreated pediatric OSA can cause high blood pressure, changes to the heart, and challenging behaviors, or even alter the childs’ growth. Unlike adults, the symptoms of childhood-onset OSA are more varied and change with developmental age which creates difficulties in both the diagnosis and patient management. Prevalence of OSA in children and adolescents is in the range of 1-5\%. It is also believed to negatively influence school performance and learning potential. Prompt treatment is a necessity, but long wait times and delays in diagnosis are overly prevalent. This is most problematic for pediatric patients as prompt treatment is recommended to prevent morbidities or the condition's exacerbation. The gold standard for diagnosis of pediatric OSA is by overnight polysomnography (PSG)  in a hospital or sleep clinic. In many countries, access to PSG is severely limited, and many children do not have confirmation of the diagnosis before treatment. Consequently, some children who do not have OSA will undergo unnecessary surgery to remove their tonsils and adenoids while other children with serious OSA will go untreated. Thus, simple and accessible options to identify children with OSA are greatly needed. 
A possible simpler approach to diagnosis than PSG might be to examine the structure of a patient's face with the goal of identifying facial features that indicate a high risk for the presence of OSA.
Face verification consists of representations of a patient's face that extract important features.  A distance measure can be used to determine the similarities and dissimilarities between pairs of faces.  Mathematically, the face features lie in a metric space, a space where the distance between two objects can be defined.  Groups of objects close together form a cluster distinct from other groups. Clusters of similar faces can be constructed in such a setting using tools from statistics and machine learning. 
One such tool is, persistent homology, which is a branch in computational topology that was recently developed (\cite{EdelsbrunnerPersistentSurvey}, \cite{Zomorodian2005ComputingHomology}).  This novel approach combines mathematical and statistical methods to the analysis of features on a surface--e.g. a face. A persistent homology analysis requires a set of filtration parameters, which will affect the performance of the method. These would need to be set manually by the researcher. 
As an alternative, the advent of applied deep learning allows us to examine novel methods for OSA face classification such as the convolutional neural network (\cite{Szegedy2014GoingConvolutions}, \cite{Szegedy2016Inception-v4Learning}, \cite{Szegedy2015RethinkingVision}) where parameters are learned automatically over many iterations of the algorithm. In this article, we compare classification of 172 facial images obtained from persistent homology (\cite{Lesnick2015InteractiveModules}), curve shape analysis (\cite{Srivastava2011ShapeSpaces}), and convolutional neural networks. The clinical study is approved by the ethical board of University of Alberta~(Pro00057638). 

\begin{figure}[htp]
	\centering
	\includegraphics[scale=0.32] {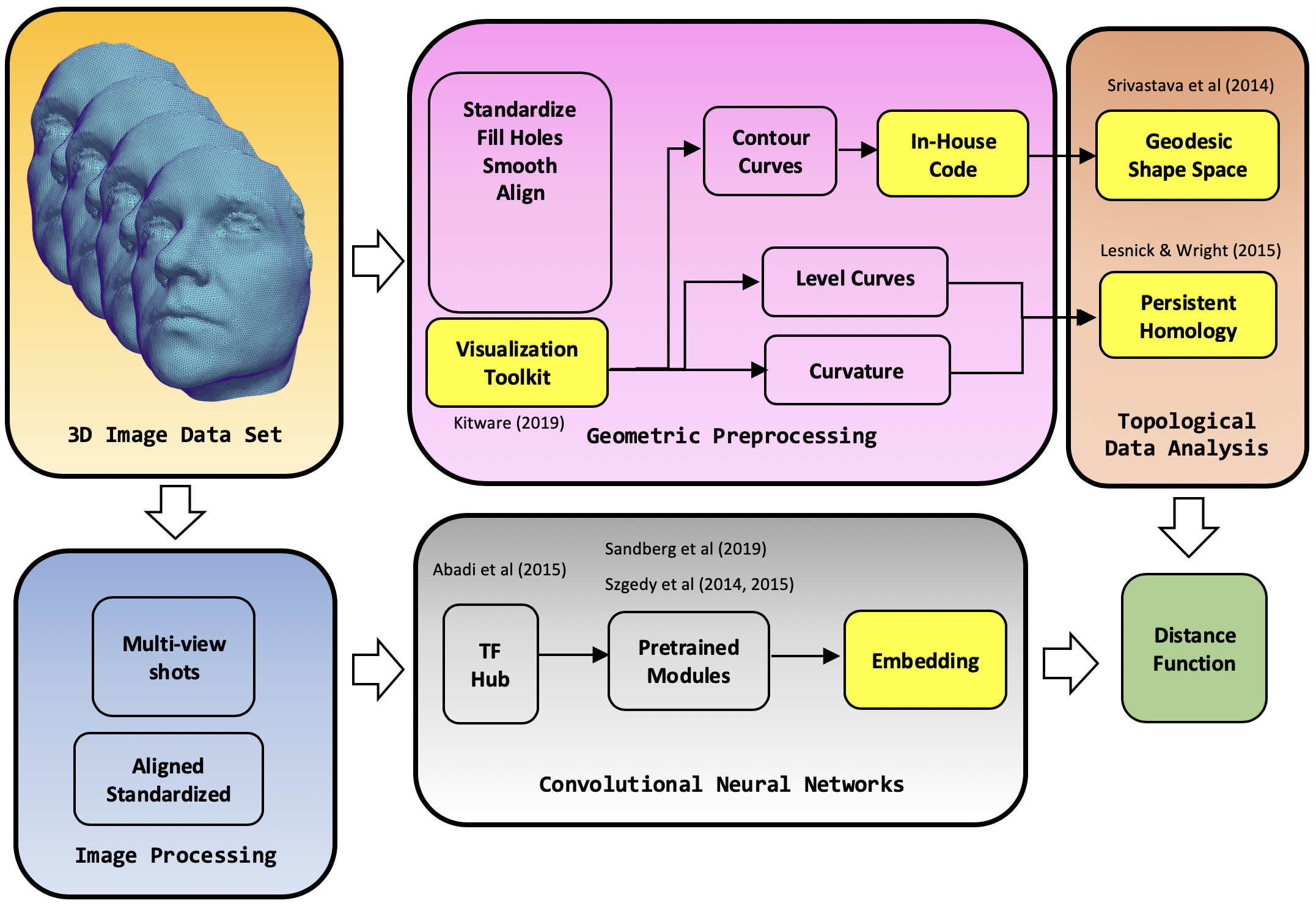}
	\caption{Flowchart of procedure followed in this article.}
	\label{fig:flowchart}
\end{figure}

We organize our works as follows, see the flowchart of this article in Figure~\ref{fig:flowchart}. 
In Section \ref{sec:contribution}, we briefly summarise computational methods on face recognition and our contribution to face image analysis. In Section~\ref{sec:methods}, we summarize persistent homology, curve shape analysis and the convolutional neural network methodology. We explain how the data was preprocessed in Section \ref{sec:processing} followed by  results in Section~\ref{sec:dataAnalysis} for two-parameter persistent homology, geometrical shape analysis, and convolutional neural network. We complete our article with conclusion in Section \ref{sec:conclusion}. 

\section{Previous Works and Our Contribution}
\label{sec:contribution}
\subsection{Previous Works}

Various approaches to computational face recognition have been explored. Pioneers in the field made progress by using heuristic operations to extract representations from a face. In the \emph{feature-based} or \emph{landmark} approach, face data is processed to find significant anatomical features such as the nose, ears, and chin and use this to reduce the dimensionality of the input data. These features include quantities such as surface area and positions of center of mass in some reference coordinate system. The history of feature based methods goes back to the work of \citet{Kanade1973PictureFaces}. More elaborated studies  were built upon this initial framework such as deformable templates (\cite{Yuille1992FeatureTemplates}), Hough transformation (\cite{Nixon1985EyeRecognition}), Reisfeld operation (\cite{Reisfeld1994GeneralizedRecognition}), and dynamic link structures (\cite{Wiskott1997FaceMatching}). Although these studies provided improved  accuracy, they all suffer from the requirement of manual work for each subject. Hence, they are not practical for a large number of subjects. \citet{Campadelli2005ACharacterization} offered an automated remedy to this problem. The feature based method falls under the so called \emph{local realization} category. 

In a global or holistic method the whole input data, an entire face image, is considered for evaluation. A  direct analysis of input data is not applicable due to the \emph{curse of dimensionality}. \citet{Sirovich1987Low-dimensionalFaces} used Principal Component Analysis (PCA) to obtain a low dimensional representation of each face. Based on this, \citet{Turk1991FaceEigenfaces} created \emph{eigenfaces} which uses eigenvector basis functions to provide an efficient method for dimension reduction. This method improved drastically in work of \citet{Pentland1994View-basedRecognition}. However, it was shown that PCA fails to identify the same face when more than one observation from same subject is present (\cite{Belhumeur1997EigenfacesProjection}). In this case, Fisher's Linear Discriminant Analysis (LDA) offered better performance \cite{Moon2015LDA-basedCooperation}. Subsequently, Independent Component Analysis (ICA) (\cite{Comon1994IndependentConcept}), a method for decomposing a signal into independent parts, is utilized by some studies to retain more complex structures in faces \cite{Low2019Multi-FoldRecognition}.

Facial or craniofacial morphology in obstructive sleep apnea is based on landmarks chosen from 2D images. Using lengths and/or angles between landmarks, multivariate analysis of variance is applied to compare facial or craniofacial shapes between OSA patients and normative subjects (\cite{Luzzi2016CraniofacialSnoring.}, \cite{Espinoza-Cuadros2015SpeechAssessment}).

In this work, we conduct a side-by-side study of deep learning and persistent homology as with the advancement of applied deep learning, many classic methods in the field of computer vision and specifically face recognition are being outperformed. 
We consider the weight-shared convolutional neural network (CNN) ( \cite{LeCun1999ObjectLearning}). This type of network was subsequently used to analyze the MNIST (\cite{LeCun1998Gradient-BasedRecognition}) data set of handwritten digits. The resulting network was extended and applied to finger print classification (\cite{Baldi1993NeuralRecognition} )
and employed for off-road obstacle detection (\cite{Lecun2005Off-RoadLearning}).

There are various famous CNN approaches dedicated for face representation and identification. Deep CNNs are proven to enhance the recognition performance in various benchmarks and contests (\cite{Szegedy2014GoingConvolutions}, \cite{Szegedy2016Inception-v4Learning}, \cite{Szegedy2015RethinkingVision}). The deeper networks have more layers to capture a wider range of features. Several challenges appear as the network gets deeper. The most familiar one is the vanishing and exploding gradient (VAEG) (\cite{Glorot2010UnderstandingNetworks}). In an attempt to address these problems, new networks were developed such as Inception and Residual networks that try to resolve the VAEG issue by adding parallel paths and offering mechanisms to bypass intermediate layers (\cite{He2015DeepRecognition}). 
DeepFace network has five convolution and two fully connected layers and was created by \citet{YanivTaigman1999DeepFace:Verification}. DeepFace is trained by four million images of four thousands individuals. VGG-Face network is another example and uses thirteen convolutional and two fully connected layers and was trained on 2.6 million face images of 2600 individuals. The most renowned work is FaceNet by \citet{Schroff2015FaceNet:Clustering}. In FaceNet, the face image is mapped to an Euclidian space where distance is the dissimilarity between two faces. In FaceNet one convolution and eleven inception modules, sixty-seven convolution layers and two fully connected layers are used. An Inception module is a combination of convolution and max pooling layers. FaceNet is trained by two hundred million face images of two million individuals.  

\subsection{Contributions}

Our objective for this article is to compare CNN classification performance to that of two competing methods: persistent homology and a differential geometric approach via shape analysis. Our CNNs are based on two main architectures. The first architecture is inspired by the work of \citet{Zeiler2014VisualizingNetworks} and uses one $1 \times 1$ convolution based on \cite{Lin2013NetworkNetwork}. The second approach is the inception network based on the work of \citet{Szegedy2015RethinkingVision} at Google and is explained in more details below. There are more recent deep learning studies for facial recognition
(\cite{Li2015ADetection},  \cite{Chen2015FacialIdentity.}, \cite{Ranjan2016AnAnalysis}) which could be basis for future investigation. 
We will apply two dimensional persistence homology techniques that are developed by (\citet{Lesnick2015InteractiveModules}). Several methods in shape analysis will be used (\cite{Grenander1993GeneralStructures}, \cite{Kendall1984ShapeSpacesb}, \cite{Younes1998ComputableSHAPES}). Furthermore, we adapt the geometric approach proposed in (\citet{Srivastava2011ShapeSpaces}). We will explore whether measurements of craniofacial form 3D-photographs improve our ability to identify children with OSA without the need for PSG whose technology is often limited or unavailable. 
The main goal of this work is to classify pediatric patients' facial images into two categories: risk of OSA or no risk of OSA.

The current study tries to utilize persistent homology and available pre-trained CNNs to analyze the face of subjects and to classify the severity of pediatric OSA based on the labeled data. In persistent homology, 3D images of faces are used. However, in deep learning, multi-angle 2D images are inputs for network. Although, utilizing the 3D model in deep learning is expected to provide more accuracy due to more available information, a standard pre-trained model for such data is still not widely accessible. The full training of a 3D face CNN requires a massive dataset and lies beyond the scope of the current study, but would be beneficial for possible future investigation.

\section{Methods}
\label{sec:methods}
\subsection{Persistent Homology}

Geometry is the study of shapes, and topology is the study of the connectivity. Consider the 26 letters of the English alphabet, $ \mathsf{A}$, \ldots, $ \mathsf{Z}$.
Geometry classifies these letters into 25 equivalence classes: $\{\mathsf {N, Z}\}$ due to their rotational symmetry, 
and the remaining 24 letters each form one class as they are all unique shapes.
In contrast, topology creates 3 groups, those with a single hole $\{\mathsf{A,D,O,P,Q,R}\}$, those with two holes $\{ \mathsf{B} \},$ and those with no holes 
$\{\mathsf{C, E ,F, G, H,I, J, K, L, M, N, S, T, U, V, W ,X,Y,Z}\}$, with each group corresponding to a homotopy equivalence class---i.e. a collection with the same number of holes.
While geometry is too fine, topology is too coarse in distinguishing between topological objects.
Persistent homology, on the other hand, 
combines the differentiating power of geometry and the classification power of topology (\cite{Collins2004AData}).

To obtain the homology groups of a manifold $X$---e.g. each alphabet letters--we first sample points randomly, $\mathcal{S}=\{X_1, \ldots, X_n\}$ 
from $X$. We then calculate homology groups of $X$ using $\mathcal{S}$, by constructing a simplical complex, which is a collection of simplices. 
Vietoris-Rips (\cite{EdelsbrunnerPersistentSurvey}) is the most well known simplicial complex and is computationally efficient. Hence, we will work with the Vietoris-Rips complex throughout this article.
Simplicies in this complex $V_t$ are formed by joining points
$x_i$ and $x_j$ if the pairwise  distance $d(x_i, x_j)$ is less than or equal to some $t$.
For example, a 1-simplex is an edge
joining two points if  $d(x_i, x_j)\leq t,$ a 2-simplex a triangle joining three points where all  pairs of points have pairwise distance less than $t$.
Then, for each dimension~(degree) $p \geq 0$, we consider the $p$-th homology group $H_p(V_t)$, which captures the information of the $p$-dimensional hole structure in the simplicial complex $V_t$.
The $p$-th Betti number ($\beta_p$) is the rank of $H_p(V_t),$ and it counts the number of $p$-dimensional holes in a complex $V_t.$
For example, a $0$th degree hole is a connected component, a $1$st degree hole is a loop, and a $2$nd degree hole is a enclosed space or void.

The complexes $V_t$ are nested, $V_t \subset V_{t'},$ for $t \leq t',$ and this inclusion induces a linear map $H_p(V_t)\rightarrow H_p(V_{t'}).$
Although the homology group can be calculated at each fixed $t$, computational topology examines the evolution of simplicial complexes and its homology groups  as $t$ increases. The images of these linear maps are \emph{persistent} homology groups.
This evolution can be examined by three different descriptors: barcodes, persistence diagrams, and persistence landscapes, which are the most commonly used topological descriptors.
Within a filtration (evolution), persistent homology captures all topological characteristics in a complete, discrete invariant.  
We consider the half-intervals denoted as $[b, d)$ where $b$ and $d$ are the times when the $p$-degree hole in the complex is formed (born) and when it is filled-in (died), respectively.  A barcode $\mathcal{B}_p(V)$ is when this invariant is expressed as a multiset of half-intervals for each degree $p \geq 0$  (\cite{Collins2004AData}).
The interval $[b, d)$ can be also presented as a  
point $(b,d)  \in \real\times(\real\cup{\infty})$.  The collection of all such points is the persistence diagram (\cite{EdelsbrunnerPersistentSurvey}). 
The stronger a topological feature is in the complex, the longer the interval. In other words, `true' feature of a complex would persist throughout the evolution process while noise will appear and disappear quickly.
The third descriptor is the persistence landscape (\cite{Bubenik2015StatisticalLandscapes}), a functional summary from which mean and standard deviation can be calculated.

At a fixed $t$,  the barcode $\mathcal{B}_p(V_t)$ is the $p$-th Betti number $\beta_p$  for the simplicial complex $V_t,$ for each degree $p.$ 
A barcode $\mathcal{B}_p(V_t)$ absorbs not only the persistence of each topological feature (hole) but also the relations and evolution history between topological features, while $\beta_p(V_t)$ simply counts the number of holes for each complex at a fixed $t$.  That is, $\mathcal{B}$ has persistence whereas $\beta$ is static.

\begin{figure}[!htbp]
	\begin{center}
		\includegraphics[scale=0.9]{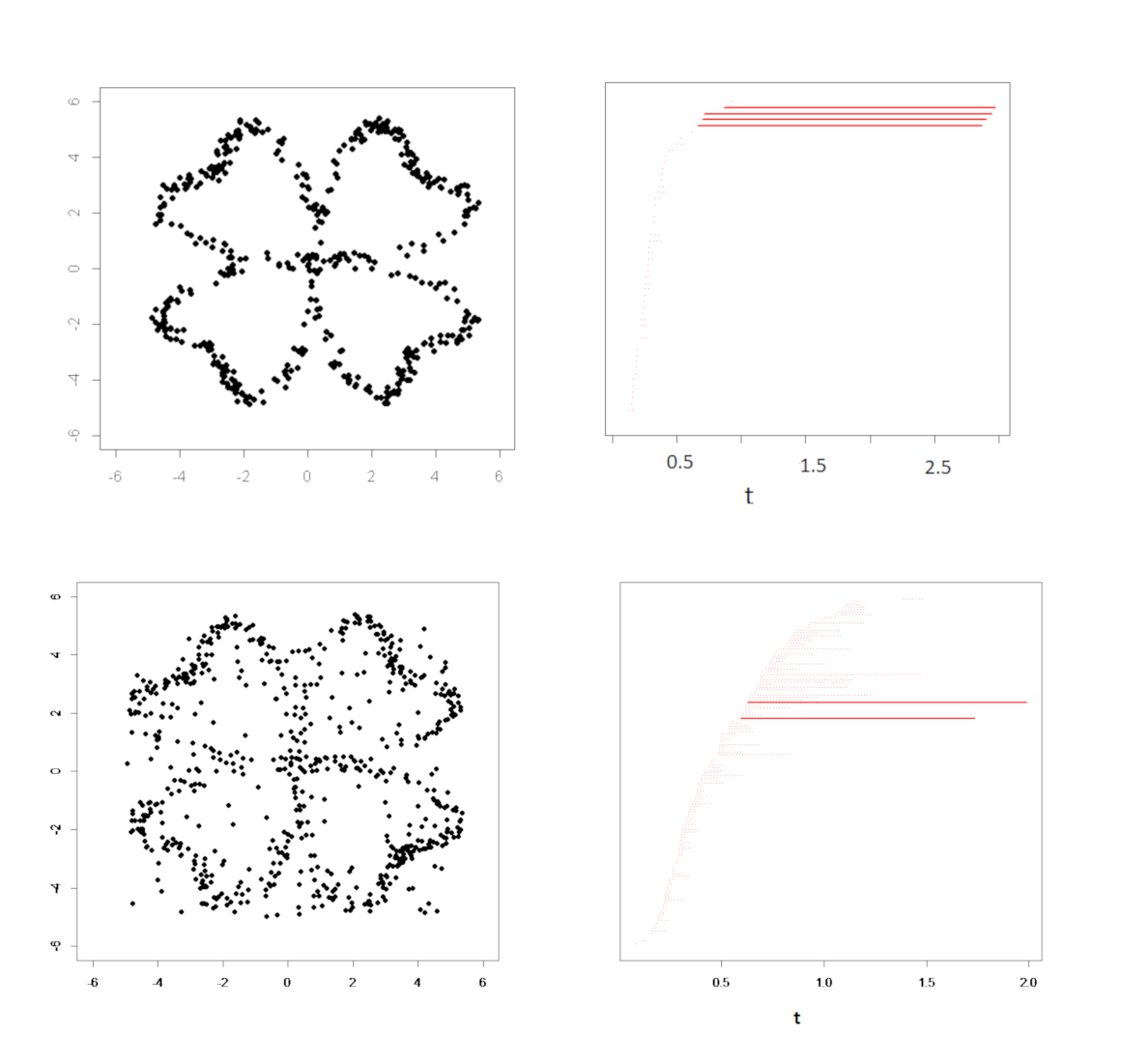}
		\label{fig:clover1}
	\end{center}
	\caption{(Top-left) The most prominent feature of a hollow four-leaf clover are 4 loops (one dimensional hole). (Top right)  The persistent interval indicates this feature in the $\beta_1$-barcode.
		(Bottom left)  Adding a few points to the middle of the leaves. (Bottom right)  Persistence of the  2 holes disappears and $\beta_1$-barcode  indicates only two loops.}
	\label{fig:clover}
\end{figure}

In Figure \ref{fig:clover}, we illustrate the persistent homology of a Vietoris-Rips complex with random points taken from hollow four-leaf clover.
The degree 1-barcode of four-leaf clover indicates four long (persistent) intervals which correspond to four loops.

\begin{figure}[htp]
	\begin{center}
		\includegraphics[scale=0.2]{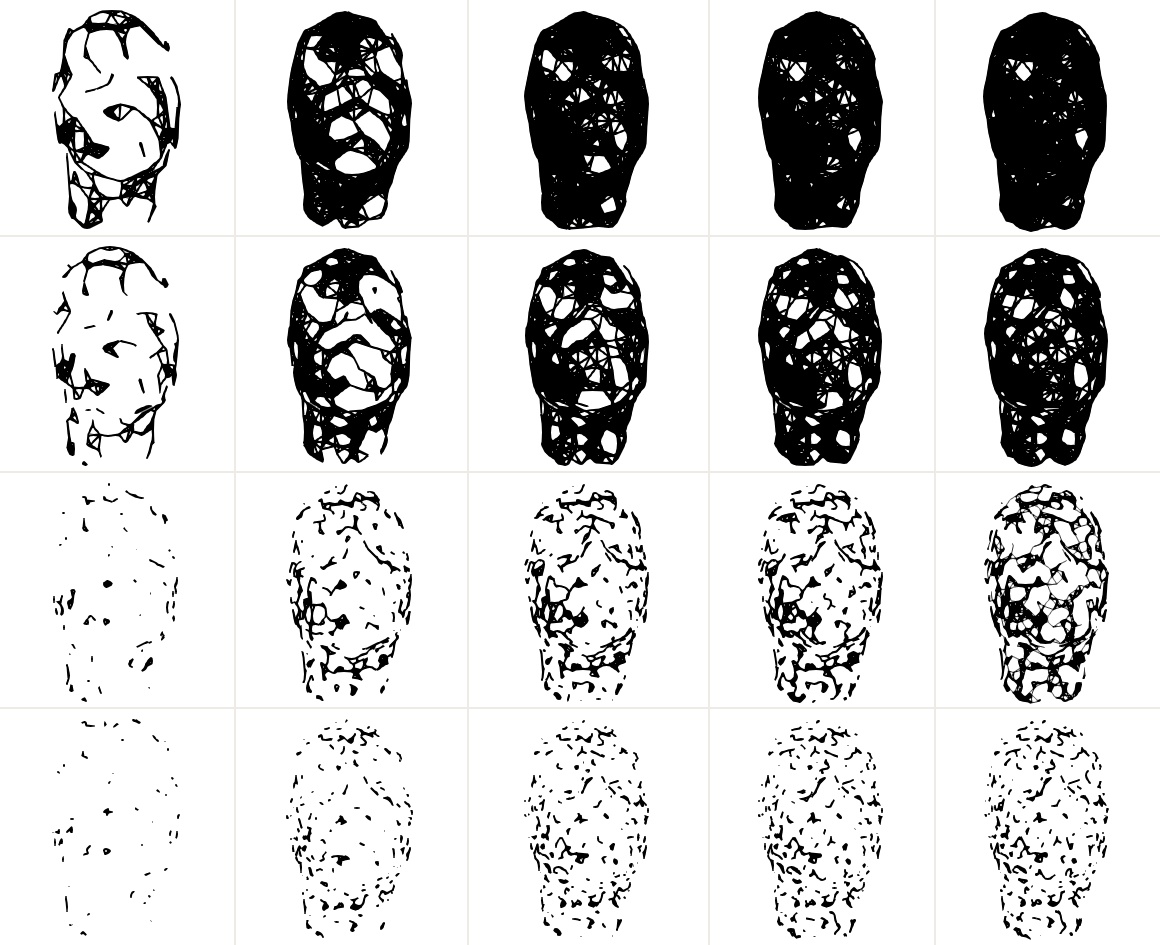}
	\end{center}
	\caption{
		Bi-filtration of point cloud of a face. Filtration of complexes both parameters $t$ in y-axis and $\kappa$ in x-axis. 
		There are more simplices as the distance $t$ between points  increases and more points as the curvature $\kappa$ at points increases. 
	}
	\label{fig:biFilt}
\end{figure}

One-parameter persistent homology~(PH), indexed by $t$, has shown to be effective in distinguishing signals from noise. Many researchers have successfully applied this to data analysis, see works of (\citet{Carlsson2009ThePersistence}, \citet{Lee2017IntegratedHomology}, \citet{Topaz2015TopologicalModels}), to name a few.
However, one-parameter PH is not robust against outliers and noise. As shown in Figure \ref{fig:clover},  $\beta_1$-barcode for hollow four-leaf clover with a few points in the middle  of loops indicates two persistent loops whereas  the `true' feature should be four loops. 
Also, one-parameter PH is not sensitive  to discern a circle (sphere)  from  an ellipse (ellipsoid), for example.
To incorporate cases with outliers and noise, it is common to  consider an alternate (or in addition to $t$) parameter such as a density estimate.
To distinguish a circle from ellipse, (\citet{Collins2004AData}) suggested a curvature~$(\kappa$) as a parameter at each fixed  inherent parameter $t$.

The most effective way to deal with two parameters is to filter complexes for both parameters simultaneously, that is, obtain multi-parameter PH. 
Figure \ref{fig:biFilt} depicts  bifiltration of face point cloud data. Similar to the one parameter case, bifiltration corresponding to 2-parameter PH is a set of  complexes that are nested $V_{t, \tau} \subset V_{t', \tau'},$ when $t  \leq t'$ and $\tau \leq \tau'.$ We will use $V$ for both single and two  parameter filtration and should be clear in the context. Bifiltration in Figure~\ref{fig:biFilt} can be written more precisely;
$V_{t, \tau}=V (f^{-1}(-\infty, \tau])_t,$ where $V$ is Vietoris-Rips complex and $f:\mathcal{S} \rightarrow \real,$ a curvature function defined on point cloud $\mathcal{S}.$ Each column in Figure \ref{fig:biFilt} shows simplices with the points whose curvature size is smaller than a fixed 
$\tau,$ then build up more simplicial complexes on only these points by increasing $t.$

A major problem is that there is no closed solution for multi-parameter PH.  More precisely, there is no corresponding complete invariant one parameter barcode in the multi-parameter setting (\cite{Zomorodian2005ComputingHomology}). 
The main difficulty for utilizing multi-parameter PH is that its representation is much more complicated. Contrary to one-parameter PH where its information can be completely captured by the barcode, multi-parameter PH does not have such a complete discrete invariant (\cite{Zomorodian2005ComputingHomology}). However, for analyzing and comparing multi-parameter PH obtained from the data, its discrete invariants are desirable even if the captured information is incomplete. One such invariant is the rank invariant (\cite{Cagliari2009One-DimensionalHomology}, \cite{Cerri2009MultidimensionalStable}), which roughly is to utilize the ranks of all the linear maps $H_{p}(V_{t, \tau}) \to H_{p}(V_{t', \tau'})$.

We will apply RIVET, the Rank Invariant Visualization and Exploration Tool, to the invariants suggested in multi-parameter PH (\cite{Lesnick2015InteractiveModules}).  
RIVET computes and visualizes  three 2D invariants;   the \emph{fibered barcode}, the \emph{dimension (Hilbert) function}, and the \emph{bigraded} Betti numbers. 
In this article, we will only consider the dimension~(Hilbert) function in the PH approach to face classification.
Let $\n$ be the non-negative integers. For each degree $p$, the $p$-th dimension~(Hilbert) function is a map $ h: \real^2 \rightarrow \n,$ such that 
$h^p_V(x):=\beta_p(V_x), $ where $x=(t, \tau).$  In other word, $h^p_V(x)$ is the $p$-th Betti number $\beta_p$  of the complex at fixed parameter 
$(t, \tau).$

\begin{figure}[!htb]
	\centering
	\includegraphics[scale=0.7]{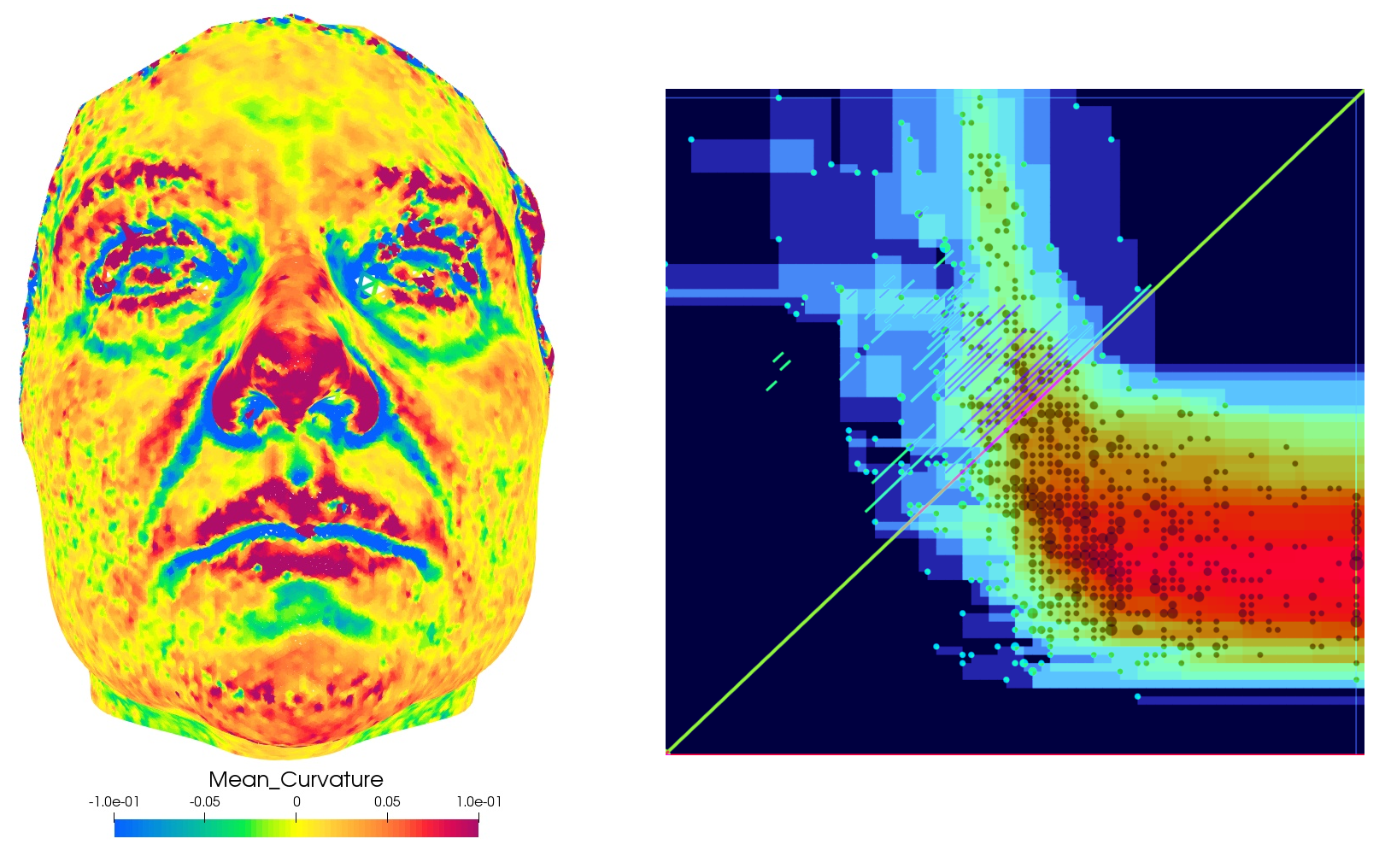} 
	\caption{(Left) Mean curvature value at each point is shown via a linear color map. Red indicates highest and blue indicates lowest curvature values. (Right) Fibered barcode and Hilbert function, and bigraded Betti numbers in degree 1 on a rectangle over the  range $[0, 40]\times [-0.5, 0.5]$ with respect to the radius and curvature. The fibered barcode that is restricted to a diagonal line is drawn in purple. The shades in cells indicate values of dimension function. Darker colours indicate higher function values. The circles show bigraded Betti numbers in degree 1.
	}
	\label{fig:FiberedHilbertF}
\end{figure}

We visualize a volunteer's face image with mean curvature values, fibered barcode, dimension (Hilbert) function, and bigraded Betti numbers in Figure~\ref{fig:FiberedHilbertF}.

\subsection{Shape Analysis of curves}\label{sec:shape}

Shape analysis in mathematics and statistics has a long history. There have been  roughly two competing approaches: landmark based
and continuous curves and surfaces based. 
For brief comparisons between two methods, see (\citet{Klassen2004AnalysisSpaces}, \cite{Srivastava2011ShapeSpaces}) and references therein.
To define a shape space precisely, tremendous amount of theoretical background in geometry is needed.  Thus, we give a brief intuitive overview.
A preshape space $\mathcal{C}$ is a collection of, for example, all closed curves.
A shape space $\mathcal{S}$ of $\mathcal{C}$ is a quotient  space $\mathcal{C}$ modulo some group $G$. That is, $\mathcal{S}$ contains sets of curves from $\mathcal{C}$ that can be transformed into each other by actions in $G$.  Therefore, a shape space is a space of equivalence classes---i.e. collections of equivalent curves up to deformation.
For our purposes,
we denote $\mathcal{S}^o$ and $\mathcal{S}^c$ to be the shape spaces corresponding to preshape spaces $\mathcal{C}^o$ and $\mathcal{C}^c$, respectively, where $\mathcal{C}^o$ is the collection of open curves and $\mathcal{C}^c$ of closed curves in $\real^d$.

For our analysis, we applied the techniques in (\cite{Srivastava2011ShapeSpaces}) and now briefly summarize their work.
They construct shape spaces of square-root velocity~(SRV) representing both open and closed curves.
That is, a parametrized curve is a map $\alpha: X\rightarrow \real^d,$ where $X=[0,1]:=I$ for open curves and $X=S^1$ for closed curves.   
We restrict to curves that are absolutely continuous on $X.$
The SRV is defined as a function $\nu:X\rightarrow \real^d,$ such that $\nu(t)=\dot{\alpha}(t)/\sqrt{\|\dot{\alpha}(t)\|},$ 
where $\|\cdot\|$ is the Euclidean norm in $\real^d.$
We note that the curve $\alpha$ can be recovered from SRV, $\alpha(t)=\int_0^t \nu(s) \|\nu(s)\| ds.$
The reason for shape analysis on SRV representation is that it makes computation of geodesics in preshape and shape spaces more efficient.
To analyze shapes of curves, we need to present curves that are invariant with respect to translation, scaling, rotation and reparameterization.
Removing scaling and translation variability results in a preshape space of curves.  If one is interested in the size~(length) of curves,  scaling transformation is not necessary.
In our data, the shapes of curves are important but not the size and so scaling variability is removed.
Further transformations of rotation and reparameterization makes preshape space into shape space. 

By imposing the constraint $\int _X \|v(t)\|^2\, dt=1,$ the scaling and translation variability are handled. 
The preshape spaces of 
open $\mathcal{C}^o$ curves and closed $\mathcal{C}^c$  curves in $\real^d$  are defined in terms of their SRV functions;
\begin{align*}
	\mathcal{C}^o&=\left\{ \nu\in \mathbb{L}^2(I, \real^d) \,\middle|\, \int _{I} \|v(t)\|^2\, dt=1\right\}     \\
	\mathcal{C}^c&=\left\{ \nu\in \mathbb{L}^2(I, \real^d) \,\middle|\, \int _{S^1} \|v(t)\|^2\,dt=1, \int _{S^1} \nu(t)\|\nu(t)\|\,dt=0\right\}   
\end{align*}
We denote $\mathcal{C}$ for either $\mathcal{C}^o$ or $\mathcal{C}^c.$ 
The rotation and reparameterization of a curve $\alpha$ are denoted by the actions of $SO(d)$ and $\Gamma$ on its SRV, where
$SO(d)$ is the special orthogonal group of $d\times d$ matrices---i.e. rotations---and $\Gamma$ is the set of all orientation-preserving diffeomorphisms of $X.$
These two actions are presented as the product of groups $SO(d)\times \Gamma$ on $\mathcal{C}$
$((O, \gamma), \nu)=O(\nu \circ \gamma)\sqrt{\dot{\gamma}}$
where 
$\nu \circ \gamma=\nu(\gamma(t))\sqrt{\dot{\gamma}(t)}.$

\begin{figure}[htp]
	\includegraphics[scale=0.03]{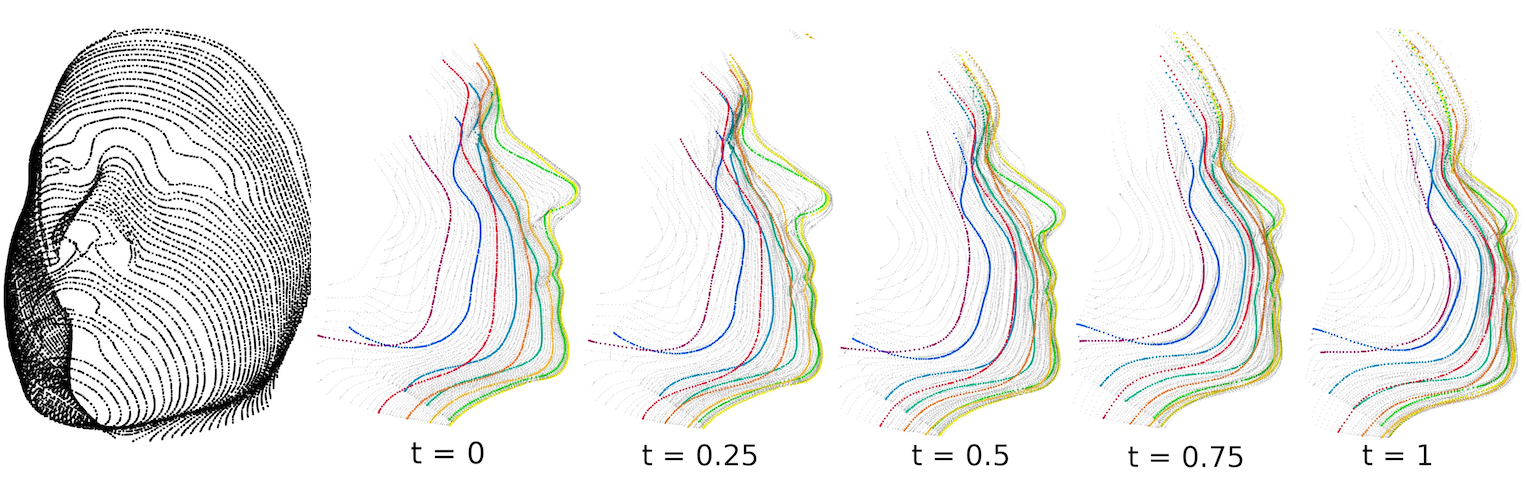}
	\caption{(First figure) Facial curves. (From second to fifth figure) Geodesics between two patients (side-view) with  few highlighted curves.}
	\label{fig:curvesNgeodesic1}
\end{figure}

We present face curves and geodesic paths between two faces in Figure \ref{fig:curvesNgeodesic1}.

\subsection{Metrics for classification and statistical inference} \label{sec:inference}
Starting from the original data, we transformed a manifold (face image) into  either a point cloud in $\real^3$ or curves in $\real^3.$
For point cloud data, we applied bi-filtration with a curvature measurement as a second parameter. 
For facial curves, similar to bi-filtration using point cloud in Figure \ref{fig:biFilt}, bi-filtration is performed to curves that are lower than each fixed height in face image (highest point on the face is a tip of nose).
We then computed dimension functions. In other words, the manifold was transformed to a topological descriptors on each data set: point clouds and curves.



For our data analysis in section~\ref{sec:dataAnalysis}, we apply a dissimilarity measure on dimension (Hilbert) functions.
We consider two such metrics: the $\ell^k$ and Hausdorff distances.
Let $h$ and $h'$ be two dimension functions. The $\ell^k$ distance is defined as $d(h, h')= (\int (h-h')^k\, dA)^{1/k}$ where $A$ is 
suitably truncated.
Hausdorff distance $d_H$  between two sets $X$ and $Y$ is defined as 
$$
d_{H}(X,Y):=\max\left\{ \sup_{x\in X}\inf_{y\in Y}\left\Vert x-y\right\Vert ,\,\sup_{y\in Y}\inf_{x\in X}\left\Vert x-y\right\Vert \right\}.
$$ 
In our analysis below, we apply $\ell^2$ and maximum of Hasudorff norm over sublevel sets, each norm is denoted by $d^e$ and $d^H.$
\begin{align}
	\label{EqnL2} d^e &:=d^e(h, h')=\left( \int (h-h')^2\, dA\right)^{1/2}  \\
	\label{EqnHaus} d^H&:=\sup_{\lambda \in \real}\left[d_H (\left\{x\in X: h(x)\leq \lambda \right\}, \left\{y\in Y: h'(y)\leq \lambda\right\}) \right].
\end{align}

For facial curves, after obtaining their preshapes and shapes, we compute geodesics between curves as described in \citet{Srivastava2011ShapeSpaces}.
We note that $\mathcal{C}$ is a submanifold, more precisely, $\mathcal{C}^c \subset \mathcal{C}^o \subset \mathbb{L}^2(X, \real^d).$ 
Therefore, restricting  the standard metric on $\mathbb{L}^2(X, \real^d)$ to the submanifold $\mathcal{C}$, $\mathcal{C}$ inherits Riemannian structure including geodesics, shortest geodesics, and geodesic length between elements in $\mathcal{C}$, which can all be obtained.
Let $\rho: I \rightarrow \mathcal{C}$ be a parameterized curve with $\rho(0)=\nu_0$ and $\rho(1)=\nu_1.$
The length of $\rho$ is defined as $L(\rho)=\int_I \|\dot{\rho}(t)\|^{1/2} \, dt,$ and $\rho$  is called a minimizing geodesic if $L(\rho)$ achieves minimum over all geodesic paths. The length of the minimizing geodesic in preshape space $\mathcal{C}$
is the geodesic distance, $d^c(\nu_0, \nu_1)=\inf_{\rho} L(\rho).$
The geodesic distance between any two points in shape space $\mathcal{S}$  is given by
\begin{eqnarray}\label{EqnGeo1}
d^s([\nu_0], [\nu_1])=\inf_{(\gamma, O)\in \Gamma\times SO(d)} d^c(\nu_0, O(\nu_1\circ \gamma)\sqrt{\dot{\gamma}}), 
\end{eqnarray}
where $[\nu]$ indicates orbit of a function $\nu \in \mathcal{C}.$ 

To compute geodesic distance between two points in shape space $\mathcal{S}^o$ or $\mathcal{S}^c$, 
we solve the joint  optimization problem in equation (\ref{EqnGeo1}).
Define the cost function, $\mathbf{C}: \Gamma\times SO(d) \rightarrow \real,$ such that $\mathbf{C}(\gamma, O)=d^c(\nu_0, O(\nu_1\circ \gamma)\sqrt{\dot{\gamma}}).$
For the open curves $\mathcal{C}^0$, the solution to the joint minimization is obtained by individual optimization over $\Gamma$ then over $SO(d).$ The optimization techniques are well known, Dynamic Programming algorithm (\cite{Frenkel2003CurveMethod}), for example.
For the closed curves $\mathcal{C}^c, $ the minimum cannot be achieved directly, but is computed by a gradient-based iterative method.
That is, group actions of $\gamma$ and $O$ are sequentially applied, $\gamma^{m}= \gamma_1 \circ \cdots \circ \gamma_m$ and
$O^{m}= O_1 \cdots O_m$ and minimize  $\mathbf{C}(\gamma^{m}, O^m)$ until the incremental change in the cost function is small.
For facial curves in our data, there are several curves on each face, we choose $k$ curves (either open or closed) at a particular resolution. 
We calculate geodesic distances, 
$d^s(\mathcal{S}_i, \mathcal{S}'_i), i=1 \ldots k,$ between  each pair of curve shapes, $\mathcal{S}_i$ 
and $\mathcal{S}'_i$ , then we combine these geodesic distances into one quantity, so geodesic distance between two facial shapes is
\begin{eqnarray}\label{EqnGeo}
d^g:=d^{g} (\mathcal{S}, \mathcal{S}')=\prod_{i=1}^{k} d^s (\mathcal{S}_i, \mathcal{S}'_i).
\end{eqnarray}

The final format of our facial images from the $n$ patients will be an $n \times n$ symmetric  matrix of pairwise distances between facial images.
Five distance matrices for dimension functions obtained from two-parameter persistence will be presented in Section 3 
corresponding to $\ell^2$ distance $d^e$ in equation (\ref{EqnL2}) and Hasudorff distance $d^H$ in Equation (\ref{EqnHaus}) with respect to both curvature and facial curves. The fifth distance matrix consists of geodesic distances $d^g$ in Equation (\ref{EqnGeo}).
Applying a permutation test with 1000 random permutations to each distance matrix, we can calculate a p-value for the null hypothesis that facial shapes of OSA and no OSA subjects have no significant difference. 

\subsection{Convolutional Neural Networks}

Persistent homology and geometric shape analysis can capture some significant structures of a face image. Those methods attempt to extract the visual OSA patterns in a face by using a combination of the curvature, geodesic distance, and level curves. 
A supervised deep feed-forward neural networks can also explore the geometry and provide insight regarding further significant structures. In contrast to persistent homology, the parameters in neural networks are not predefined manually, but are instead weights based on the network architecture.

\begin{figure}[ht]
	\centering
	\includegraphics[scale=0.2] {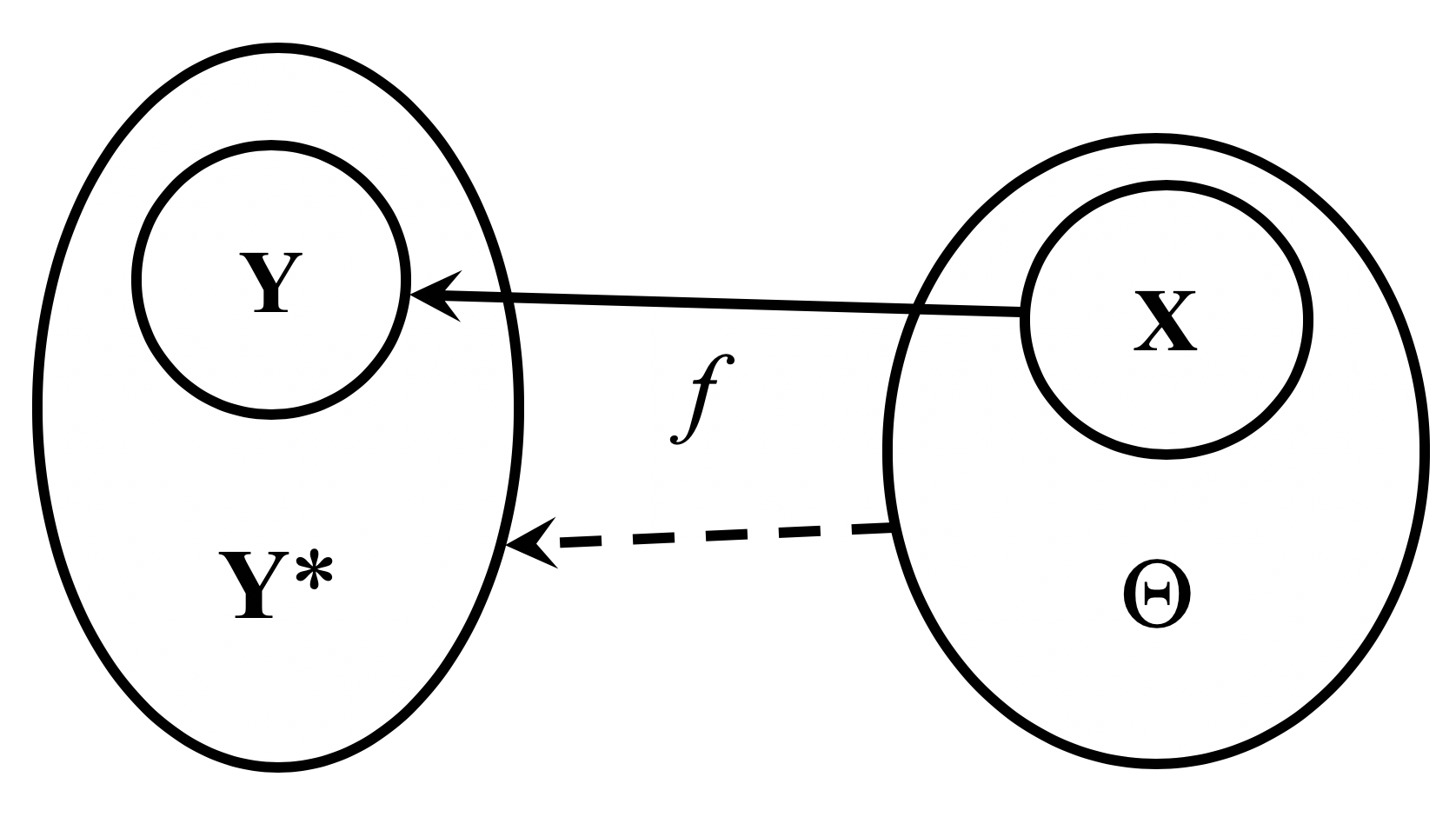}
	\caption{Map diagram to describe a neural network as a function. Here, function $f$, which is found by mapping training data to known representation (solid arrow), is used to map a larger space which contains the training data (dashed line). }
	\label{fig:neuralmap}
\end{figure}

Neural networks can be considered as a large\footnote{Large in scale or complex in its functionality.} nonlinear function $f: \mathbb{R}^{p \times n}  \xrightarrow{} \mathbb{R}^m$ depending on some weight matrix $\textbf{W}$ which take $n$ images $\textbf{X} \in  \mathbb{R}^{p \times n}$ as input and provide a representation $\textbf{Y} \in \mathbb{R}^m$  as output. The weight matrix $ \textbf{W} \in \mathbb{R}^{m \times n}$ is a learned parameter of $f$. When the function is found---i.e. $\textbf{W}$ is computed---it can be used to estimate a larger space---i.e. a matrix with more data---by the map $f: \Theta \xrightarrow{} \textbf{Y}^{*} $, in which $\Theta$ and $\textbf{Y}^{*}$ are its domain and range spaces. This map can be viewed in Fig \ref{fig:neuralmap}.


Let us assume that a network consists of multiple layers. Each layer is labeled with an index $ i \in \{ 1, \dots, M \} $.
The function $f$ can be defined as composition of functions $\textbf{A}^{(i)}(\textbf{W}^{(i)T}\circ)$ in which $\textbf{A}^{(i)}$ is a nonlinear operator\footnote{Within domain of neural networks, this type of operator is called activation function. Various types of activation functions are examined in this study. e.g. RELU, tanh, and sigmoid.}, $\textbf{W}^{(i)T}$ is the transpose of weights matrix and $\circ$ is the operation within the network. 

Thus, in the recursive form an $M$-layer network can be formulated as
$\textbf{H}^{(i)} = \textbf{A}^{(i)} (\textbf{W}^{(i)T} \circ \textbf{H}^{(i-1)})$ for $i \in \{1,\ldots,M\}$
in which $\textbf{H}^{(i)}$ is called a hidden layer and belongs to a certain vector space. The learning task is to find weights (kernels) by minimizing a cost function $\mathcal {L}$. The cost function is usually defined as the distance between known outputs $\textbf{Y}$ and unknown outputs $ {proj \textbf{Y}^{*}} _{\textbf{Y}} $ in a metric space. The optimization problem for a weight matrix can be written as
\begin{equation}
\textbf{W}^{*} = \argmin_{\textbf{W}}  {\mathcal {L}}(\textbf{Y}, {proj \textbf{Y}^{*}} _{\textbf{Y}})
\end{equation}

The network output should provide a representation of the input space by embedding it into another space. An embedding is expected to preserve the important topological properties of the input. This has a crucial consequence in performance of network with respect to the learning tasks. It is common to assume that the input is defined in a reference metric space. This means for all vectors $\textbf{x}, \textbf{y}$ and $\textbf{z} \in \Theta$  it could be written
\begin{align}
	\label{eqn:eqnmetric}
	L d_{\Theta} (\textbf{x}, \textbf{y}) \leq d_{\textbf{Y}^ *} (f(\textbf{x}), f(\textbf{y})) \leq LC d_{\Theta} (\textbf{x}, \textbf{y}) 
\end{align}
in which $d$ is a measure for distance. Furthermore, $L$ is the scale and $C > 0$ is the distortion where both are bounded constants.  Hence, the embedding map is 
$(\textbf{X}, d_{\Theta}) \xhookrightarrow [\text{}]{\text{C}} (\textbf{Y}^*, d_{\textbf{Y}^*})$
.
In classification, the embedding provides metrics that are coinciding with the input metric $d_{\Theta}$. An embedding output would be desirable for a classification task as they reduce the dimensionality of the input. This is necessary for efficient evaluation. Another important consequence is that, embeddings makes notion of clustering. Hence, it is a crucial part of the classifier.

\subsubsection{Network Specifications}

Historically, CNNs are well-established for two dimensional multi-channel images classification. Although these established frameworks could be extended to directly address the 3D data as it was implemented in PH, a multi-view shot approach is another convenient approach to assess a three dimensional individual subject face.
Various CNN architectures are examined in this study including InceptionV3, InceptionResV2, MobileNetV2, and NASNetLarge.  Each network is developed with a specific goal but all are proved to be powerful for image classification. 

We focus our detailed examination on the network called InceptionV3, which consists of \textit{Inception} modules which repeat consistently throughout the network \cite{Szegedy2015RethinkingVision}. Inception networks follow four criteria: (1) they try to avoid bottleneck close to the representation layer; (2) they are fast to train. This is a result of \emph{feature disentanglement} by large convolutional 
factorization; (3) they use \emph{spatial aggregation}, which leads to non-significant loss; (4) Width and depth of network are balanced to avoid the well-known VAEG in backpropagation process.

Each Inception module contains several parallel paths. Several convolutions exist within a path and have kernels of sizes $1 \times 1$, $3 \times 3$ and $5 \times 5$. Larger kernels from previous versions of Inception are decomposed into a number of smaller kernels which are convoluted sequentially. There exist layers with  $1 \times n$ and $n \times 1$ kernels and pooling layers, which are used to reduce the dimensionality.

Let us reuse the formulation used by \citet{Szegedy2015RethinkingVision} on developing the Inception V3 network. The notations here are consistent with this research article. It is assumed that the network calculates a notion of probability of the outcome as the \emph{soft-max} operation on logits $z_k$, given the ground truth label are $k \in \left \{ 1, ..., K \right \}$ . For training $x$, it can be written as
$p (k|x) = e^{z_k}(\sum_{i=1} ^K e^{z_i})^{-1}$.


The logits themselves are calculated for the output of the network. Assuming normalized ground truth (OSA image labels) to be $q_k$,
the loss function is calculated according to the cross entropy and the training process tries to optimize parameters by minimizing this function. For the training purpose, label distribution can be replaced by 

\begin{equation}
q' (k|x) = (1 - \epsilon) q(k|x) + \epsilon u(k)
\end{equation}

in which $u(k)$ is uniform distribution \footnote{The uniformity condition is not necessary. However, the distribution is considered to be fixed.} and  probability $\epsilon \in (0,1)$. For the simple case that $q_k=\delta_{k,y}$, where $\delta$ is the Dirac delta, the ground truth can be smoothed and called \emph{label-smoothing regularization}. This can be written as 
\begin{equation}
H(q', p)=-\sum_{k=1}^{K} q'_k \log p_k = (1-\epsilon)H(q, p) + \epsilon H(u, p)
\end{equation}
where $H$ is cross entropy loss function. In this context $H(u, p)$ is the difference between $p$ and some fixed distribution $u$. 

This is similar to the method of \citet{Zeiler2013StochasticNetworks} for regularization with penalty. In this method, an interpolated cost is to be minimized. This ensures that a portion of the artificial accuracy is to be removed from the training procedure. 

A challenging obstacle in front of current learning tasks would be small sample sizes for training data with respect to the complexity of the classification task. Consequently, the model is extremely prone to \emph{over-fitting}.
A few tricks are employed to remedy over-fitting. One is to dropout some of the components randomly based on the work of (\citet{Srivastava2014Dropout:Overfitting}). Dropout reduces the estimation bias towards the training data and forces the  realization of extensive patterns instead of ``memorizing'' the details or noise in training images. 
Furthermore, a way to improve the fit is augmentation in which training data is manipulated to produce a larger dataset.  In this study several random distortions of image properties including random crop, scale, brightness and horizontal flip are implemented. The following has been used to help with avoiding the overfitting and accuracy: batch normalization; early stopping; and larger batch size.

\subsubsection{Transfer Learning}

Training the CNN model from scratch (random initialization) for specific task of OSA face image classification is outside of scope of this study. The computational solution of such network would be extremely expensive. Moreover, finding a suitable starting point for these parameters would be difficult. The computational resource problem becomes more evident as the state of the art models grow deeper, wider and more complex in architecture. Furthermore, the training task requires large number of pediatric OSA patients images. Lastly, supervised labeling of such large image library would not be trivial due to the significant manual labour of expert staff.

Training the randomly initialized CNN is not only impractical but it is even unnecessary. A pretrained network\footnote{In this study tensorflow-hub is used to obtain models which are pretrained on ImageNet database. TF.Hub is a library containing many useful machine learning features including deep learning pretrained models of most well established architectures. ImageNet data set \cite{NIPS2012_4824} is standard collection of labeled images which are used as both training and validation benchmark tool in many previous studies.  Knowledge from ImageNet is transferred for learning OSA faces. The concept of transfer learning has a close resemblance with inductive reasoning.
	
	The network graph information is loaded from the TensorFlow hub (TF.Hub). This is achieved by restoring the pre-calculated weights and network topology of the so called \textit{module}. In practice, the transfer learning in TF is accomplished by extending the existing python code, provided by TF and Keras official websites. Furthermore, PyTorch APIs is examined to find weather significant improvement in result could be observed. } can be used for weight initialization. Transfer learning tries to learn a new task by building upon the pre-learned tasks. For instance, a network that recognizes faces can be further extended to learn to recognize facial gestures. This process is much faster than building a task-specific network from scratch. 
Pre-trained networks can be fine-tuned \footnote{From a certain layer toward the representation layer.} by relatively small perturbations of weights. A fully connected layer is added at the top of the network and is fully trained for the classification task.

TensorFlow-gpu v1.9 is used for creating and solving the neural networks (\cite{TensorFlow}). Our instance of TensorFlow is containerized using Docker\footnote{As in Docker official website, "A container is a standard unit of software that packages up code and all its dependencies so the application runs quickly and reliably from one computing environment to another". A platform like Docker offers vast possibilities and is used for many different purposes. The topic is outside of the focus of this study. However, Docker aided with isolation of software packages and satisfied most dependencies within, while avoiding the requirement for up to date libraries (e.g. CUDA) from host operating system which could risk an unstable host. Taking this risk seems undesirable for out local machine but it is absolutely unacceptable for cloud computing servers. } which itself uses an NVidia docker container at the run time in order to use an NVidia TitanX GPU. 

\section{Data Preprocessing}
\label{sec:processing}


Our data set consists of 3D photo from 172 subjects. All patients are labeled by clinicians into categories of (1) risk of OSA or (2) no  risk of OSA. The 3D photos are represented by a triangulated mesh in stereo-lithography (STL) format. This mesh is resampled and preprocessed using certain filters---e.g. smoothing, and hole filling. The mesh improvement is accomplished while attention is paid to avoid loss of the relevant topological features of the surface. The boundaries of the mesh are unified between subjects to represent similar clips of the surface and, the coordinate system is transformed via a standardization procedure. Specifically, the origin is translated to the middle of the path between two predefined location on two eyes of a certain subject face. Then, it is rotated to match certain unifying face features---e.g. principal components of face surface.

The 3D photos can be directly analyzed within the persistent homology and geodesic framework. 
In contrast, for the CNN approach, the data is a collection of 2D color images. Images are taken using the visualization toolkit (VTK) to capture multiple side-angles pictures from 3D facial images. By considering the standardized coordinate system, the angles are kept consistent between subjects. Furthermore, the face in each image is aligned using the Multi-Task-CNN method \cite{Zhang2016JointNetworks}. All images are JPEG decoded and standardized to a specific size to satisfy the network model requirements. For instance the inception network accepts a $299 \times 299$ resolution images and NASNet accepts $331 \times 331$ images. The data is split in a roughly 80-20 $\%$ manner for training and test sets. Each subject is only present in one of the two sets.

Due to ethical constraints, no identifiable representation of these subjects can be shown in this article. At any point if a facial representation of a subject is required, the face of an adult volunteer is used instead.

The surface geometry is represented by triangulation. Curvature is calculated at each point by using the neighbouring triangles. Hence, it is invariant with respect to the global face orientation. The face surface is not closed. Thus, a valid boundary condition for the curvature is necessary. A low-pass filter is applied to the curvature to assure that the curvature value remains within a certain threshold. Figure \ref{fig:FiberedHilbertF} shows the calculation of the mean local curvature on an adult volunteer. 

As we will realize in PH section, calculations of RIVET tool can be expensive. In order to keep these calculations feasible, points are down-sampled using vtkMaskPoints filter. This filter reduces the number of points while maintaining the topological structures of the remnants of points using the spatially stratified random sample method.  

Curves depend on the choice of the reference plane. Based on this and in order to keep curves consistent through subjects, an alignment procedure is employed. For each subject, ID of three points are saved. These points include qualitative selection of the tip of the nose and an ID for each eye. In some computer vision applications this approach is called \emph{landmark-based}.

Following the previous acquisition, faces are transformed \footnote{Transformation means a combination of translations and rotations in $\real^3$. } to maintain a consistent face orientation. 
In order to extract these curves, a VTK filter called \emph{vtkContourFilter} is used. Contour filter acts on a scalar field $S(t)$ and returns edges and points with specific value for the scalar $S(t_i)=S_i$ and $ C(t) = F(S(t)) ~~ t \in \real $.

Various scalar function are examined e.g. sections perpendicular to principal axis of the surface. Based on the resulting curves, ear-to-ear contours are observed to generate the most consistent results. The software which is responsible for generating these curves, starts from one side of the face, incrementally checks the quality of the extracted curve and continues to find the next curve on the path until some stopping criteria is reached. 

In order to keep the corresponding curves between subjects comparable, certain quality checks are required including: 1) keeping the cutting mechanism consistent; 2) aligning based on eyes and nose locations; 3) Laplacian smoothing of the surface triangles; 4) reordering the curve-points to achieve clockwise spatial sequence; and 5) repopulating empty regions with points to keep the spatial distribution homogeneous.

Initially, the point numbering of an individual contour curve is based on the original triangulated surface. Hence, the point IDs appear to be in an arbitrary order. A renumbering method, finds points as a sequence and orders them in ab in-plane clock-wise manner. Level curves are resampled by using the combination of VTK methods: \emph{vtkStripper}, \emph{vtkCardinalSpline} and \emph{vtkSplineFilter} to create an equidistant grid while preserving the shape of the curve. 

\section{Results}
\label{sec:dataAnalysis}
\subsection{Persistent Homology and Geometrical Shape Analysis}

\begin{figure}[!htb]
	\centering
	\includegraphics[width=1.0\textwidth] {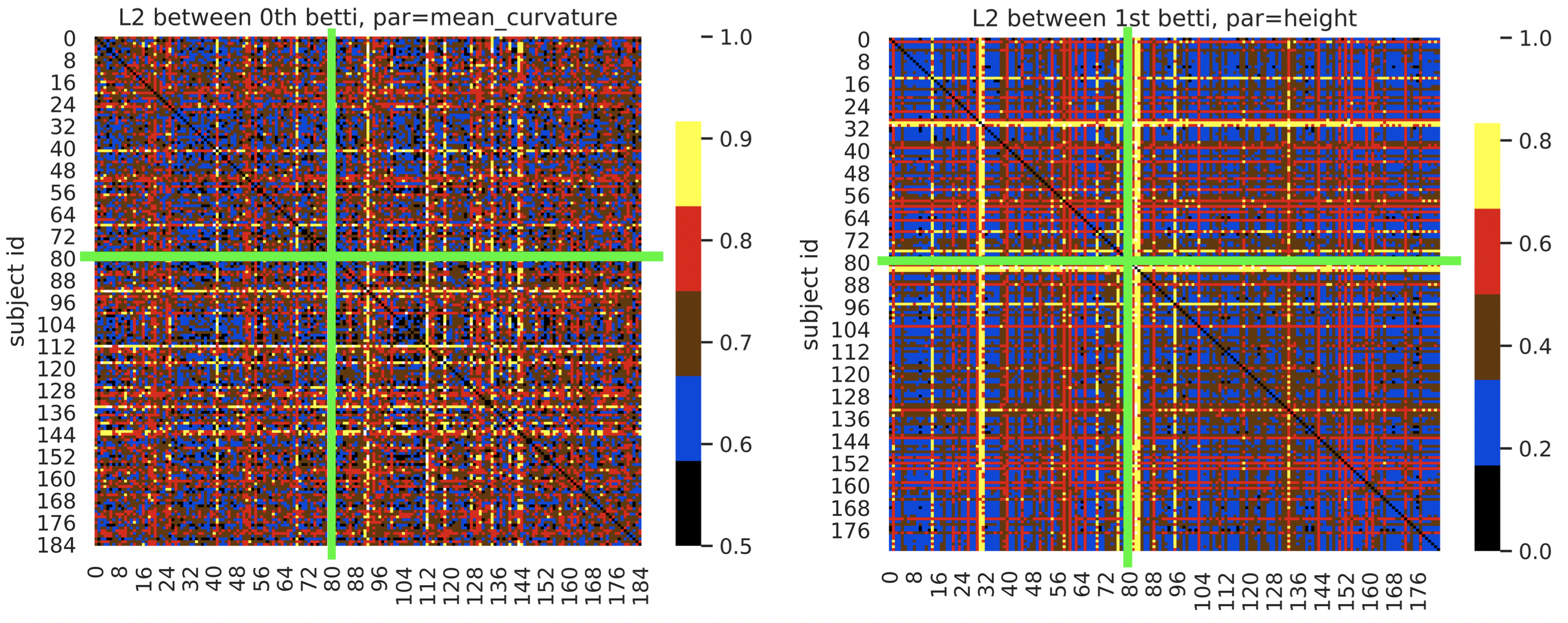}
	\caption{ $d^e$ distance matrices for two cases: curvature (left) and curves (right) are shown. Each subject's ID is enumerated from 0 to 184 (with 13 missing subjects), which denoted in vertical and horizontal axes. Subjects 0-80 are evaluated by clinicians as no risk of  OSA, subjects 81-184 risk of OSA. The thick light green lines in the middle separates patients with no OSA patients with OSA patients. As it can be observed the distance is higher on  OSA group (right) side of the line. Various Betti numbers are explored. Among those, for curvature, $\beta_0$ and, for curves, $\beta_1$ resulted in a more `structured' (see explanation in the texts) distance matrices.}
	\label{fig:distmatrix}
\end{figure}

We summarize our findings from two-parameter PH to point cloud and curves and geometrical shape analysis.
After obtaining dimension functions~(degree 0 and 1) for each 3D face image, we calculate $d^e$ (Equation \ref{EqnL2}) and $d^H$ (Equation \ref{EqnHaus})  distances between pair of dimension functions. For all 172 images, thus we have $172 \times 172$ distance matrix with respect to $d^e$ or $d^H$.
Figure \ref{fig:distmatrix} shows $d^e$ distance matrices based on two-parameter PH analysis for curvature and curves. The $d^H$ and $d^g$ (Equation \ref{EqnGeo}) distance matrices are not shown because they  are similar to $d^e$  distance matrices.

From looking at the distance matrices in Figure \ref{fig:distmatrix}, one could conclude that the curvature distance matrix shows more classification indications than the level curve distance matrix. For instance it could be argued that the distance within the children with risk of OSA  (lower right diagonal square) is larger than those among subjects with no risk of OSA (upper left diagonal square). On the other hand the distance values between subjects with no risk of  OSA and subjects with risk of OSA (symmetric off-diagonal rectangles) lies somewhere between the values for within and between the two classes (main diagonal square). Conversely the distance calculated for no risk of OSA subjects seems smaller and more homogeneous. 

\begin{figure}[!htb]
	\centering
	\includegraphics[scale=0.4] {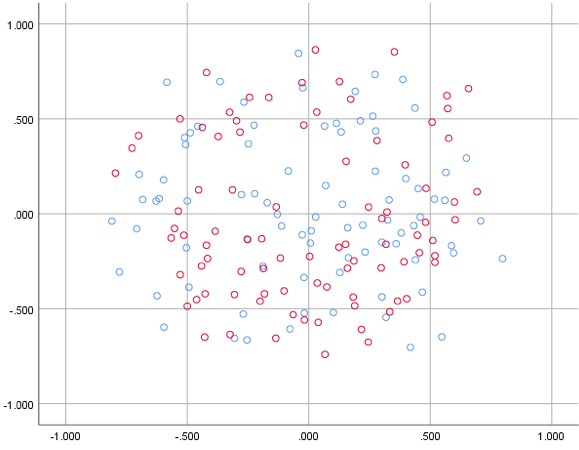}
	\caption{Scatter plot using multidimensional scaling based on $d^e$ distance. Blue circles denote subjects with no risk of OSA while red circles are patients with risk of OSA. Points live in high dimension, but plotted in two-dimensional space, where $x$-axis ($y$-axis) corresponds to the first (second) principal coordinates. (Scatter plots based on $d^g$ and $d^H$ are similar to this plot and so not presented.)}
	\label{fig:curvesNgeodesic2}
\end{figure}

However, from the distance matrices, facial shapes of risk of  OSA and no risk of OSA group are not discernible. By applying multidimensional scaling to $d^e$ distance matrix, the data is projected into two dimensions, and one can see that the subjects are not distinguishable between two groups as displayed in Figure~\ref{fig:curvesNgeodesic2}. 
A permutation test carried out using $\ell^2$ distance $d^e$ and geodesic distance $d^g$ between facial shapes. The p-values for permutation tests are 0.55 and 0.68 respectively, which is consistent with the pattern that shown in the scatter plot, Figure \ref{fig:curvesNgeodesic2}.

\subsection{Deep Learning}

\begin{figure}[!htb]
	\centering
	\includegraphics[width=0.49\textwidth] {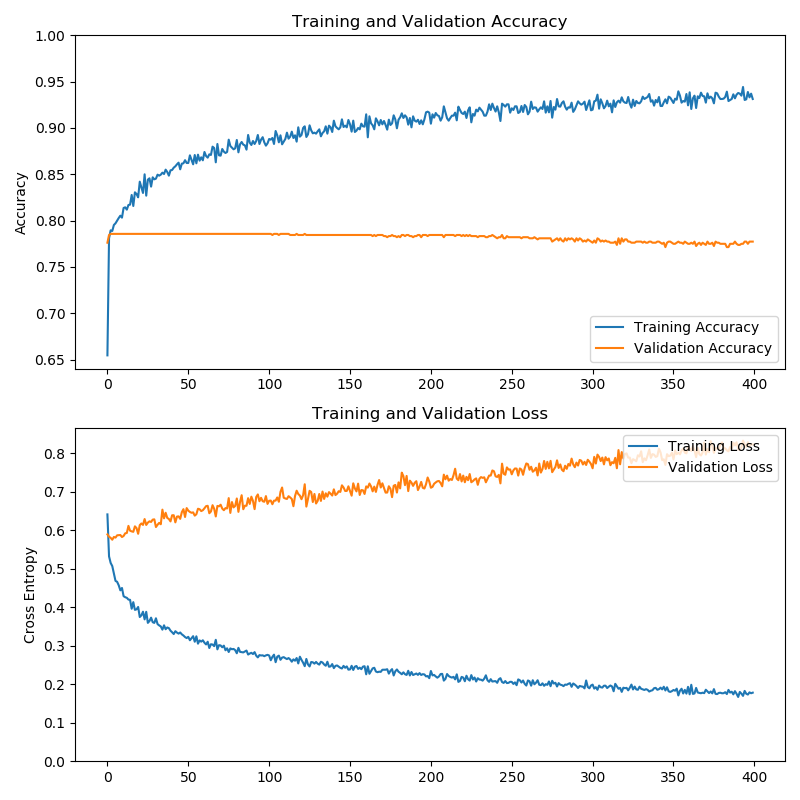}
	\includegraphics[width=0.49\textwidth] {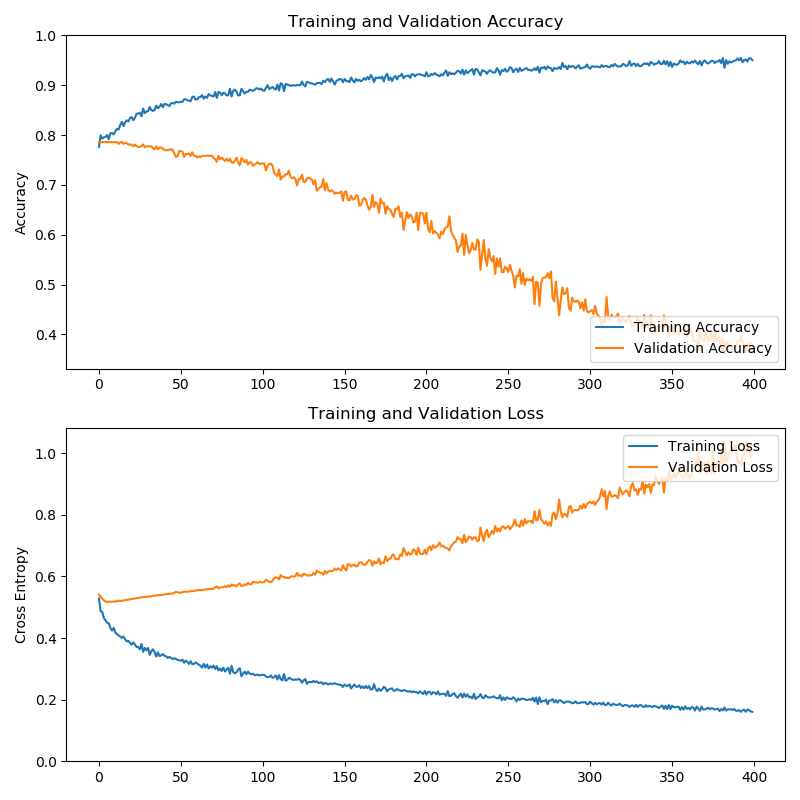}
	\caption{Iterative learning (training and validation) graphs are shown for MobileNet (left) and NasNetLarge (right). Bottom row shows the cross entropy loss while the top row shows the accuracy.  }
	\label{fig:nasnet}
\end{figure}

In order to grasp a three dimensional sense from the face surface of each subject, multi-angle 2D shots are taken from them. This is accomplished by a codes written in Python and C++ that utilize VTK libraries. All networks are pretrained by ImageNet dataset.  
The pretrained networks was unable to keep up the validation with the accuracy of the fit. Two of these training processes are shown in Figure~\ref{fig:nasnet}.

NasNetLarge contains $89M$ parameters and is much larger than MobileNet which contains $4M$ parameters. Transfer learning process is performed by fine tuning of the pretrained layers and training of an additional fully connected layer at the representation layer. 

Since the dataset for OSA training is quite limited, a large network like NasNetLarge is extremely prone to overfitting. Hence, diverging behaviore of the validation curve as it is seen in Figure~\ref{fig:nasnet} is not surprising. In such cases the network is not fine-tuned and the pretrained weight are kept frozen. 
As a result, we compare the performance of transfer learning in each network as depicted in Table~\ref{tab:somenetworks}. All training and validation sets are kept similar and are chosen from the OSA dataset. 

\begin{table}[!htb]
	\centering
	\caption{A selection of available pretrained models with training data and available platforms.}
	\begin{tabular}{|c c c c c|} 
		\hline
		-- & Architecture & Trained with & Library & Validation \\ [0.5ex] 
		\hline\hline
		1   & InceptionV3    & ImageNet & TensorFlow & $ 70 \pm 4$ \\
		\hline
		2 &   InceptionResV2  & ImageNet  & TensorFlow & $ 71 \pm 5$ \\
		\hline
		3 &   MobileNetV2  & ImageNet & TensorFlow  & $ 64 \pm 5$ \\
		\hline
		4 &   NASNetLarge  & ImageNet & TensorFlow  & $ 72 \pm 2$ \\  [1ex] 
		\hline
	\end{tabular}    
	\label{tab:somenetworks}
\end{table}

Pretrained networks are expected to handle a new task which is close enough to the original training data. However, the same expectation would not be realistic when the amount of additional data does not match the complexity gap\footnote{Higher complexity of data could be interpreted as to possess more topological structures. It is obvious but worth mentioning that having a topological structure is only definable when constraints or rules for being a structures are presupposed.} between the base learning task and the new to be learned task. 

A discussion regarding the availability of such pretrained networks seems to be necessary. As it was mentioned previously, the amount of freely available pre-labelled OSA face data in medical science research is limited. The main reason is the lack of effective communication between institutes that in turn does not improve by imposing privacy protection mechanisms. This is more problematic when a precise identifier such as human face is involved in the topic. 

\section{Conclusion}
\label{sec:conclusion}

In this article, we have examined two broad approaches towards classification of pediatric sleep apnea face: topological data analysis (TDA) and the current state of the art convolutional neural networks.  Within the topology domain, we have tried both persistent homology and geodesic distance for objects on a manifold.

Although some facial features can be associated with high OSA risk, the current TDA methods did not result in accurate classification of OSA. 
This is likely due to the fact that OSA is not an entirely anatomical problem  and OSA does not always affect craniofacial structures.

Methods from deep learning---specifically, convolutional neural networks---have shown great success when trained 
for a variety of tasks on image data.  For OSA, we tested a multitude of architectures and all resulted in over-fitting to the training set with poor performance on the validation set.  This occurred even after including many safeguards against over-fitting.  The main culprit of this failure is a lack of an adequate sample on which to train the networks. Given access to a larger sample of data, the already proven power of these methods will surely be seen.

Future efforts must emphasize improvements to deep learning algorithms in order to address medical diagnostic problems on small clinical datasets. 
Furthermore, mutually shared network models for specific diagnoses is crucial. This can be accomplished by construction of a framework where public research institutes could share their task specific pretrained models with other public institutes to further improve the models.
Specifically, when face data is involved as in OSA, the aforementioned approach would settle the challenges of ethics and privacy and provide a vast range of trained model to be used in the clinical setting throughout the world.

\section{Acknowledgements}

The authors would like to thank the National Sciences and Engineering Research Council of Canada (NSERC DG 2016-05167),
Seed grant from Women and Children's Health Research Institute, Biomedical Research Award from American Association of Orthodontists Foundation, and the McIntyre Memorial fund from the School of Dentistry at the University of Alberta.
We are grateful for support by research coordinators, postdoctoral fellows, residents and technicians in Stollery Children's Hospital and Orthodontic Clinic at University of Alberta. We thank participating clinicians and orthodontists: T. Carlyle, O. Dalci, A. Darendeliler, C. Flores-Mir, I. Kornerup, M. Lagravere, J. MacLean, P. Major, A. Montpetit, B. Pliska, and S. Quo. We thank Professor A. Srivastava for sharing Matlab code.

\bibliographystyle{model1-num-names}
\bibliography{references.bib}

\end{document}